\pdfoutput=1

\documentclass[11pt,a4paper]{article}
\usepackage[]{acl}
\usepackage{placeins}

\usepackage{times}
\usepackage{latexsym}

\usepackage{microtype}

\usepackage{booktabs, multirow} 
\usepackage{soul}
\usepackage{xcolor} 
\usepackage{multirow}
\usepackage{hyperref}
\usepackage{booktabs}
\usepackage[normalem]{ulem} 
\usepackage[utf8]{inputenc}
\usepackage[T1]{fontenc}
\usepackage[noabbrev,capitalize]{cleveref}
\usepackage[disable]{todonotes}

\usepackage{enumitem}

\usepackage{booktabs, multirow} 
\usepackage{soul}
\usepackage{changepage,threeparttable} 
 \definecolor{darkgreen}{rgb}{0.0, 0.7, 0.1}

\title{CUNI Submission in WMT22 General Task}

\author{Josef Jon  \and Martin Popel  \and Ondřej Bojar \\
       Charles University \\
  \texttt{surname@mail.ufal.mff.cuni.cz}  }
\date{}

\begin{document}
\maketitle
\begin{abstract}

\end{abstract}

We present the CUNI-Bergamot submission for the WMT22 General translation task. We compete in English$\rightarrow$Czech direction. Our submission further explores block backtranslation techniques.  Compared to the previous work, we measure performance in terms of COMET score and named entities translation accuracy. We evaluate performance of MBR decoding compared to traditional mixed backtranslation training and we show a possible synergy when using both of the techniques simultaneously. The results show that both approaches are effective means of improving translation quality and they yield even better results when combined.
\section{Introduction}
\label{sec:introduction}
    This work focuses on exploring of two methods used in NMT in order to improve translation quality: backtranslation and Minimum Bayes Risk decoding using neural-based evaluation metric as the utility function. The methods used and related work are presented in  Section 1. In  Section 2 we describe our experimental setting and results.
\section{Methods}
We describe methods we used to build our system in this section.
\subsection{Block backtranslation}
The translation quality of NMT depends heavily on the amount of parallel training data. It has been shown that the authentic bilingual data can be partially supplemented by synthetically parallel, machine translated monolingual text \cite{bojar-tamchyna-2011-improving,sennrich-etal-2016-improving,xie-etal-2018-noising,edunov-etal-2018-understanding}. Often, the synthetic and authentic parallel data are  mixed in the training dataset but previous research shows that simply mixing the two types of text does not yield optimal translation quality. We use block backtranslation (\textit{block-BT}) in  configuration similar to \citet{popel-etal-2020-cubbitt}.  This method creates blocks of parallel and synthetic data and presents them to the neural network separately, switching between the two types during the training.
Since in last year's WMT, the submission using block-BT by \citet{gebauer-etal-2021-cuni} did not find any improvements, presumably due to improperly chosen block size, we decided to verify effectiveness of this method once again.

\paragraph{Averaging type}
Previous work on \textit{block-BT} shows the importance of averaging the checkpoints to combine information from different blocks of training data in order to obtain good performance. We compare checkpoint averaging with another method of combining older sets of model's parameters with the current one -- \textit{exponential smoothing}. After each update $u$, the current parameters $\Theta_u$ are averaged (with smoothing factor $\alpha$) with parameters after the previous update $\Theta_{u-1}$: $$\Theta_{u}=\alpha\Theta_{u}+(1-\alpha)\Theta_{u-1} $$
Previous work by \citet{Popel2018MachineTU} contains experiments with  exponential averaging, but only on the level of already saved checkpoints, not online during the training after each update as for our work.

\paragraph{Minimum Bayes Risk decoding}

NMT models predict conditional probability distribution over translation hypotheses given a source sentence. To select the most probable translation under the model (mode of the model's distribution), an approximation of MAP (\textit{maximum-a-posteriori}) decoding is used, most commonly the beam search \cite{beam_search}. However, beam search and MAP decoding in general have many shortcomings described in recent work \cite{stahlberg-byrne-2019-nmt,meister-etal-2020-beam} and other approaches have been proposed to generate a high-quality hypothesis from the model.

One of them, MBR (Minimum Bayes Risk)  decoding \cite{GOEL2000115,kumar-byrne-2004-minimum}, has been proposed as an alternative to MAP. MBR does not produce a translation with the highest probability, but a translation with the best value of utility function. This utility function is usually an automatic machine translation evaluation metric. However, to optimize towards the best utility function value, it would necessary to know the ideal selection of hypothesis. In the case of MT, that would mean a perfect, best possible translation, which of course is not known during the translation process. For this reason, an approximation of the ideal translation is used, based on the model's probability distribution \cite{eikema_mbr}. This can be implemented as generating a list of hypotheses (e.g. using sampling or beam search) and then computing utility function of each hypothesis using all the other hypotheses as the ideal translation approximation (i.e. as references).
This approximation of MBR decoding can be seen as consensus decoding -- the hypothesis most similar to all  the others is chosen. Also, in this implementation, it is more appropriate to name the process reranking, rather than decoding, and we will do so from now on. 

Even though MBR is able to optimize towards many metrics and increase the scores, these gains did not translate into better human evaluation of the final translations, when using traditional metrics based on surface similarities like BLEU. Recent successes in development of novel metrics for machine translation has renewed interest in this method. \cite{comet_mbr,bleurt_mbr,muller-sennrich-2021-understanding}.



\section{Experiments}
In this section we present our experimental setup and results.
\subsection{Tools}


We tokenize the text into subwords using FactoredSegmenter\footnote{\url{https://github.com/microsoft/factored-segmenter}} and SentencePiece~\cite{kudo-richardson-2018-sentencepiece}. We use MarianNMT~\cite{junczys-dowmunt-etal-2018-marian-fast} to train the models. BLEU scores are computed using SacreBLEU~\cite{post-2018-call}, for COMET scores~\cite{rei-etal-2020-comet} we use the original implementation.\footnote{\url{https://github.com/Unbabel/COMET}}

\subsection{Datasets}
We train English-Czech NMT models for our experiments.
We train our models on CzEng 2.0~\citep{kocmi-2020-announcing}. We use all 3 subsets of CzEng corpus: the originally parallel part, which we call \textit{auth}, Czech monolingual data translated into English using MT (\textit{csmono}) and English monolingual data translated into Czech using MT (\textit{enmono}). We use \texttt{newstest2020}~\cite{barrault-etal-2020-findings} as our dev set and \texttt{newstest2021}~\cite{akhbardeh-etal-2021-findings} as our test set. 

For experiments concerning translation of named entities (NE), we used a test set originally designed for Czech NLG in restaurant industry domain\cite{dusek-jurcicek-2019-neural}.\footnote{\url{https://github.com/UFAL-DSG/cs_restaurant_dataset}} It contains sentences which include names of restaurants and addresses in Czech and their translations in English. We will call this test set the \texttt{restaurant} test set. 

\subsection{Models}
We train Transformer-base (which we denote \textit{base}) and Transformer-big (\textit{big 6-6}) models with standard parameters~\cite{vaswani-2017-attention} as pre-configured in MarianNMT. For the largest model (\textit{big 12-6}), we use Transformer-big with 12 encoder layers and depth scaled initialization \cite{junczys-dowmunt-2019-microsoft,zhang-etal-2019-improving}.\footnote{Training scripts available at: \url{https://github.com/cepin19/wmt22_general}} We also used learning rate of $1\mathrm{e}{-4}$ for the 12 layer model instead of $3\mathrm{e}{-4}$, which was used for other models.
We trained all models for at least 1.4M updates. We computed validation BLEU scores every 5k updates and we stopped if the score did not improve for 30 consecutive validations.
\todo[]{HW, GPUs, training times, base vs. big vs deeper}
We trained the models on a heterogenous grid server, which includes combinations of Quadro RTX 5000, GeForce GTX 1080 Ti, RTX A4000 and GeForce RTX 3090 cards. Typical training time on 4 1080Ti of the base models for 1.4M updates was 7 days.

\subsection{Block-BT settings}
For all our experiments, we save a checkpoint every 5k updates and we vary only the size of the blocks during which the training data stay in the same type (20k, 40k, 80k and 160k updates).  The length of the blocks is the same for all block types. We circle through the block types in the following order: \textit{auth}$\rightarrow$\textit{csmono}$\rightarrow$\textit{auth}$\rightarrow$\textit{enmono}. 

For checkpoint averaging, we average consecutive 8 checkpoints.
For exponential smoothing, we use default Marian configuration ($\alpha=0.001$, but there are some slight modifications based on number of updates since start of the training and batch size).

We also look at the effects of using only backtranslation, or both back- and forward-translation.

\begin{figure*}[ht]
\centering
\includegraphics[width=0.95650\textwidth]{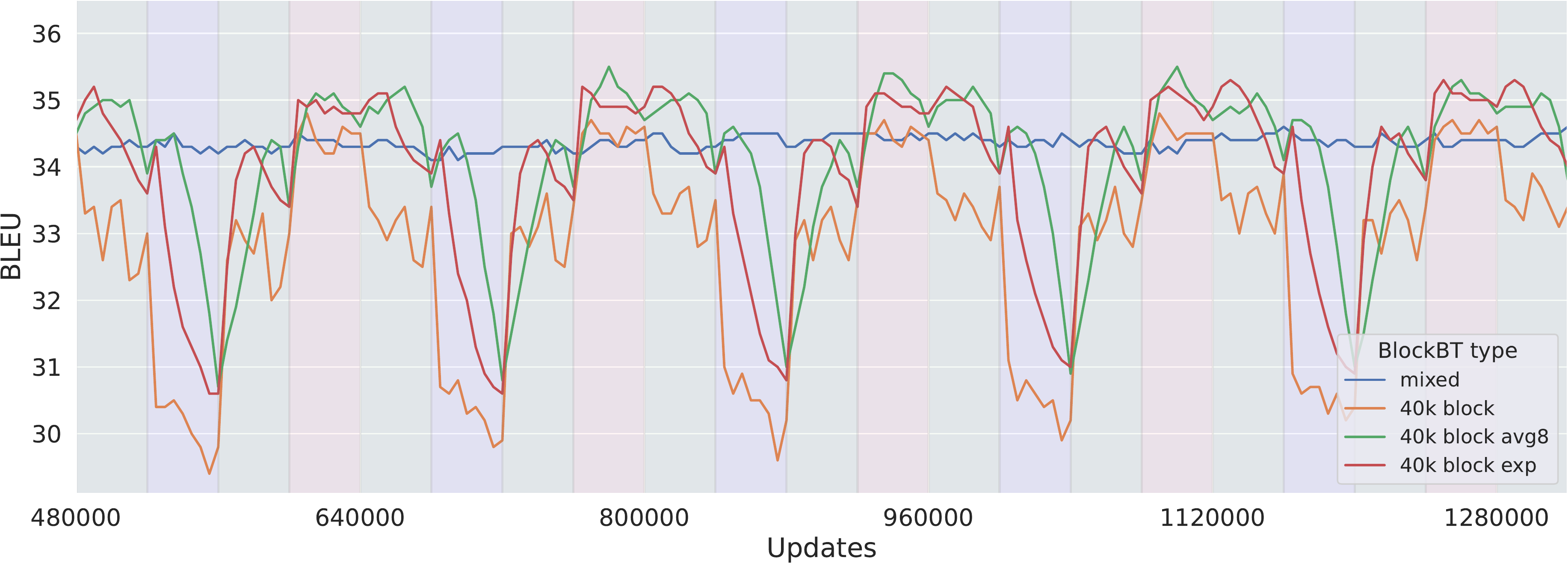}
\hspace*{-5pt}\includegraphics[width=0.95650\textwidth]{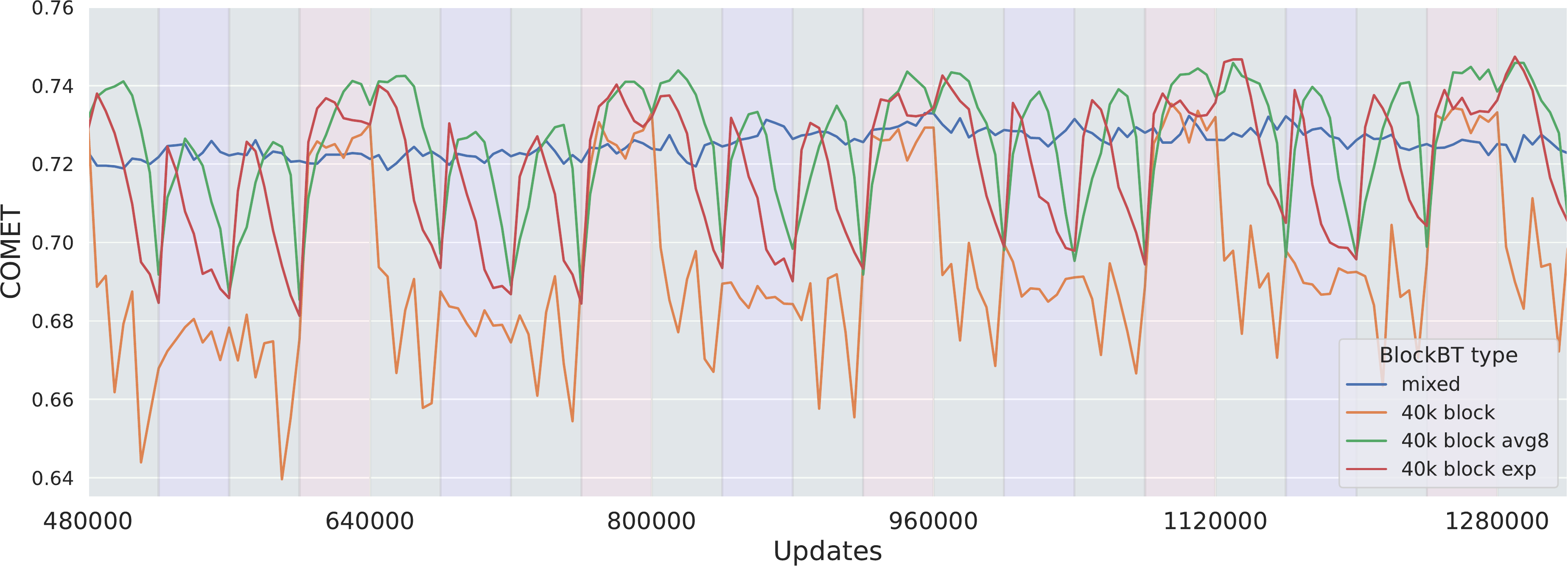}
\caption{Comparison of different training regimes for EN$\rightarrow$CS translation on newstest20 in terms of BLEU (top) and COMET (bottom). Background colors for block-BT regime show which part of training data was used for a given part of the training. Green means authentic parallel data, blue is CS$\rightarrow$EN backtranslation and red is EN$\rightarrow$CS forward translation.}
\label{fig:mixed_avg_exp_no_comb}
\end{figure*}

\subsection{Block-BT results}

\begin{table*}[!htp]\centering
\small
\begin{tabular}{lrrrrrrr}\toprule
\textbf{Model size} &\textbf{Block size} &\textbf{Avg type} &\textbf{update (k)} &\textbf{ BLEU} &\textbf{update (k)} &\textbf{ COMET} \\\midrule
\multirow{18}{*}{base} &mixed &exp &1340 &34.7 &1760 &0.7337 \\
&mixed &exp+avg8 &1365 &34.7 &965 &0.7326 \\
  \cmidrule{2-7}
&\multirow{4}{*}{20k} &- &1360 &34.6 &640 &0.7324 \\
& &exp &410 &34.9 &725 &0.7406 \\
& &avg8 &660 &34.8 &1385 &0.7349 \\
& &exp+avg8 &420 &34.9 &735 &0.7399 \\ 
  \cmidrule{2-7}

&\multirow{4}{*}{40k} &- &610 &34.8 &1415 &0.7363 \\
& &exp &1130 &35.3 &1290 &\textbf{0.7474} \\
& &avg8 &780 &35.5 &1420 &0.7462 \\
& &exp+avg8 &1150 &35.5 &1075 &0.7466 \\
  \cmidrule{2-7}
&\multirow{4}{*}{80k} &- &1250 &34.9 &960 &0.7393 \\
& &exp &1210 &35.2 &1450 &0.7447 \\
& &avg8 &985 &35.5 &665 &\textbf{0.7474} \\
& &exp+avg8 &585 &35.3 &1150 &0.7455 \\   \cmidrule{2-7}
&\multirow{4}{*}{160k} &- &1130 &34.9 &1210 &0.7387 \\
& &exp &1125 &35.3 &1285 &0.7453 \\
& &avg8 &1135 &35.5 &1305 &0.7467 \\
& &exp+avg8 &1145 &35.3 &1310 &\textbf{0.7473}  \\ \midrule
\multirow{2}{*}{big 6-6} & \multirow{2}{*}{40k} &exp &445 &35.4 &1125 &0.7546 \\
& &exp+avg8 &300 &35.4 &1310 &0.7567 \\ \midrule
\multirow{1}{*}{big 12-6}  &\multirow{1}{*}{40k}  &exp & 130 &36.1 & 1210 & 0.7848 \\ 
\bottomrule
\end{tabular}
\caption{Best COMET and BLEU scores on EN-CS newstest2020 for all the combinations of models size, training regime and block size. We report the best score and an number of updates after which was this score reached.}
\label{tab:big_results}
\end{table*}

\paragraph{Training regime and averaging method} First, we compare different training regimes: \textit{mixed-BT}, where all the training datasets are concatenated and shuffled together and \textit{block-BT} with 40k updates long blocks and two possible averaging types -- exponential smoothing (\textit{exp}) or checkpoint averaging (\textit{avg8}).

Figure \ref{fig:mixed_avg_exp_no_comb} shows the behavior of BLEU and COMET scores on \texttt{newstest2020} during the training for these configurations inthe interval between 480k and 1280k updates. The behaviour is not stable earlier than 480k steps and 1280k is the nearest lower multiplicative for the largest block size. \textit{40k block} curve represents the model without any averaging, \textit{40k block avg8} is the model trained without exponential smoothing but each checkpoint was averaged with 7 previous checkpoints for the evaluation, \textit{40k block exp} model was trained with continuous exponential smoothing. Finally,  we also experimented with combination of both --  exponential smoothing and averaging after the training. The combination does not improve over the separate averaging techniques and we omitted the curve from the figure for clarity.

In both metrics, \textit{block-BT} with either form of averaging outperforms \textit{mixed-BT} training. Without any averaging, the advantage of \textit{block-BT} over \textit{mixed-BT} is smaller. Type of averaging does not seem to play a large role -- checkpoint averaging, exponential smoothing and their combination yield very similar best scores. The best scores on \texttt{newstest2020} for each combination of parameters are presented in Table~\ref{tab:big_results}. 

The curves for checkpoint averaging and exponential smoothing behave similarly, with exponential averaging reacting faster to change of the block.  Additionally, the \textit{avg8} models have higher peaks in \textit{enmono} (red) blocks, especially for BLEU scores. 
The shape of the curves could be tuned by changing frequency of saving checkpoints and number of checkpoints to be averaged for checkpoint averaging method, or by changing the $\alpha$ factor for exponential smoothing.

There are differences in behaviour between BLEU and COMET score curves. Most notably, COMET is less sensitive to transition from \textit{auth} (green) to \textit{csmono} (blue) blocks. We hypothesize this is caused by lower sensitivity of COMET score to wrong translation of NE and rare words \cite{comet_mbr}. We present further experiments in this direction later. \todo[]{There are also peaks in forward translation, especially for BLEU and avg8 they seem higher than in auth regions, investingate in NE part.}

\begin{figure*}[ht!]
\centering
\includegraphics[width=0.95650\textwidth]{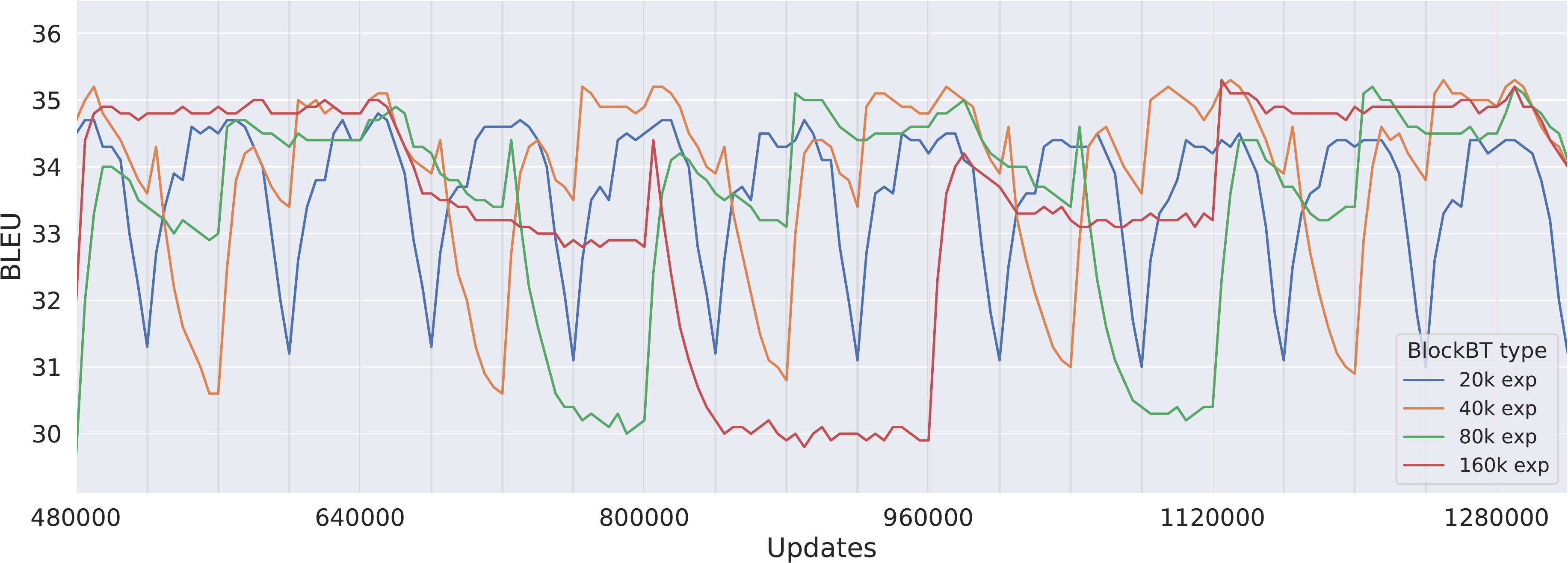}
\hspace*{-5pt}\includegraphics[width=0.95650\textwidth]{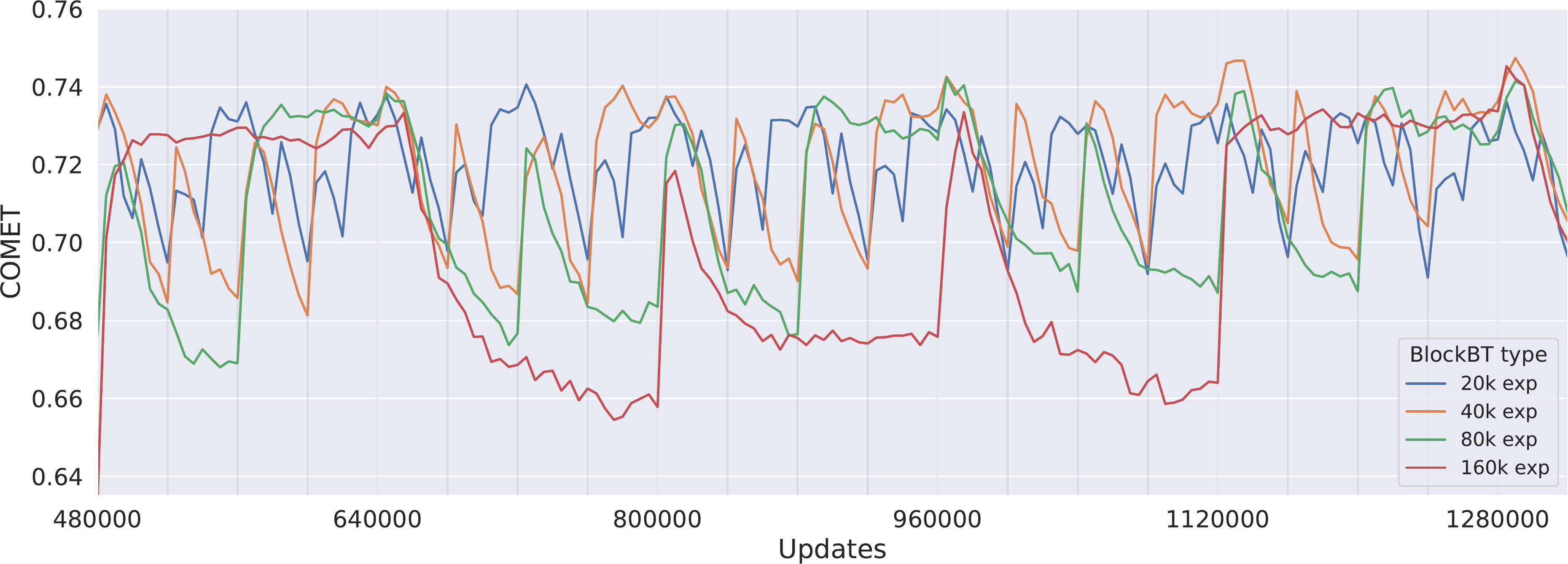}

\caption{Comparison of how the block size affects behavior of BLEU (top) and COMET (bottom) scores during the training for block-BT with exponential smoothing of the parameters, without checkpoint averaging, on EN-CS \texttt{newstest2020}.}
\label{fig:comet_block_size_exp}
\end{figure*}

\begin{figure*}[h]
\centering
\includegraphics[width=0.95650\textwidth]{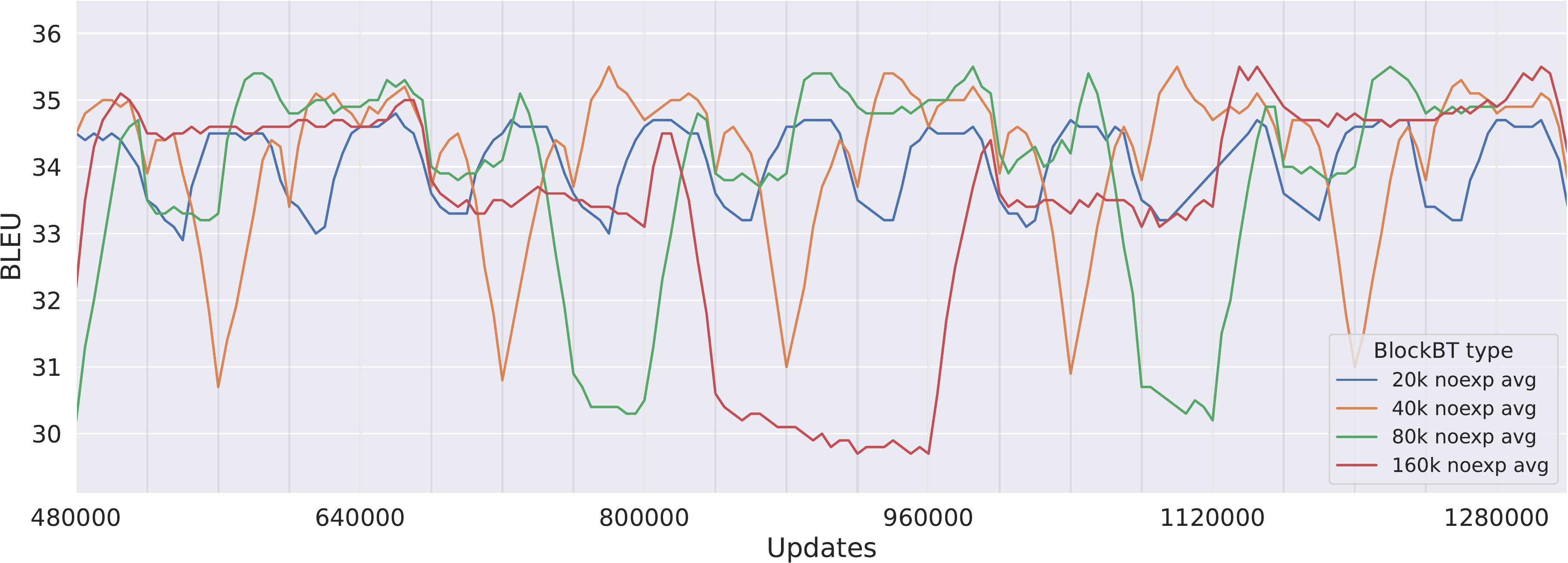}
\hspace*{-5pt}\includegraphics[width=0.9565\textwidth]{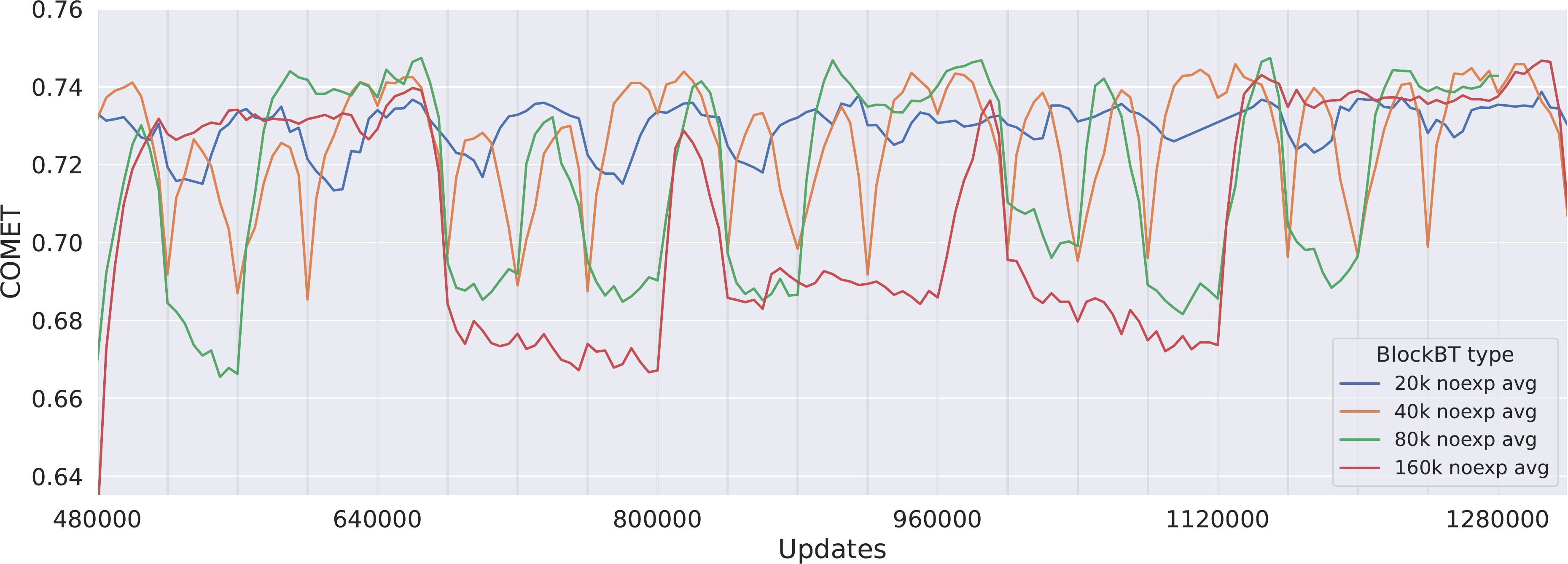}

\caption{Comparison of how the block size affects behavior of BLEU (top) and COMET (bottom) scores during the training or block-BT with checkpoint averaging and no exponential smoothing of the parameters, on EN-CS \texttt{newstest2020}.}
\label{fig:comet_block_size_noexp_avg}
\end{figure*}

\paragraph{Block size} We assess the influence of block size for both of the two averaging methods. We compare block sizes of 20k, 40k, 80k and 160k updates. The behaviour of COMET and BLEU scores is presented in Figures \ref{fig:comet_block_size_exp} and \ref{fig:comet_block_size_noexp_avg}  for exponential smoothing and checkpoint averaging, respectively. The best scores are again shown in Table \ref{tab:big_results}.

We see that 20k block size yields noticeably worse results when using checkpoint averaging
than the other sizes. The negative effect of the small block size is less pronounced when using exponential smoothing, yet still present. Other block sizes perform similarly in both metrics. This result is expected, since for 8-checkpoint averaging with 5k updates checkpointing interval, it is necessary to have a block size of at least 40k updates to fit all the 8 checkpoints and thus explore all possible ratios of \textit{auth} and \textit{mono} data.
\todo[]{So it seems that it is important to be able to fit at least n checkpoints inside a single block for n-checkpoint averaging}

\begin{figure*}[h]
\centering
\includegraphics[width=0.95650\textwidth]{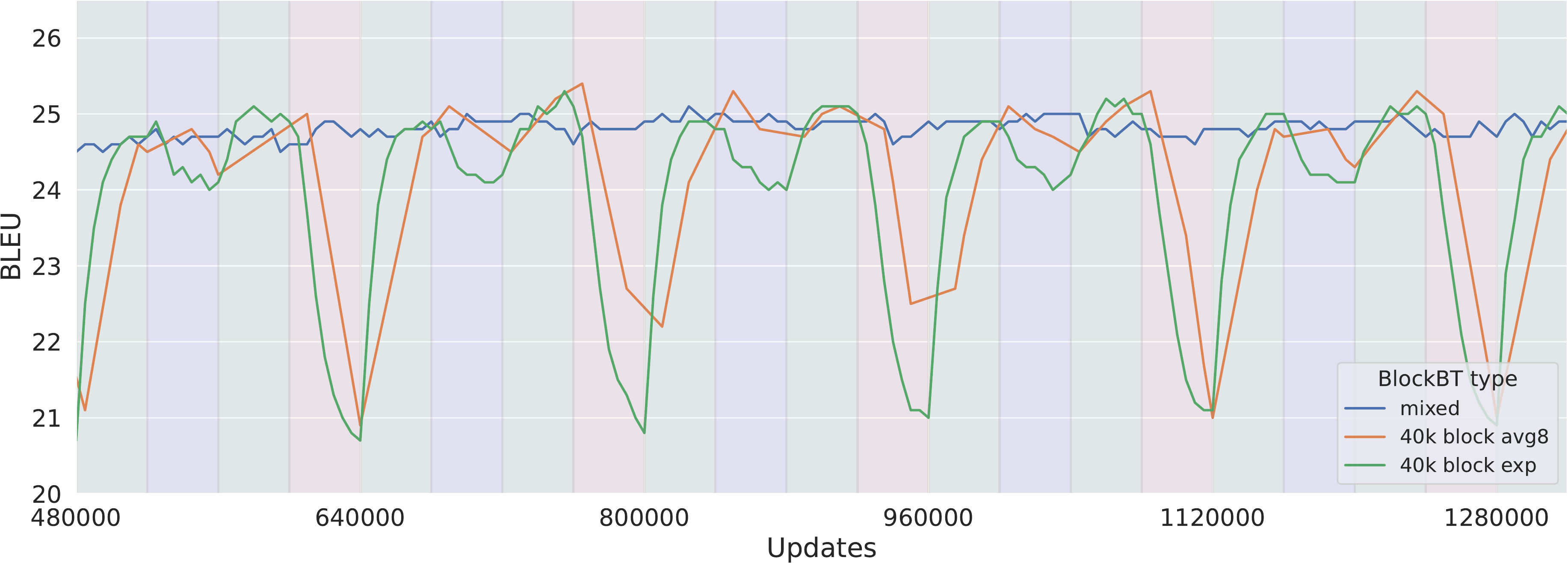}
\hspace*{-5pt}\includegraphics[width=0.9565\textwidth]{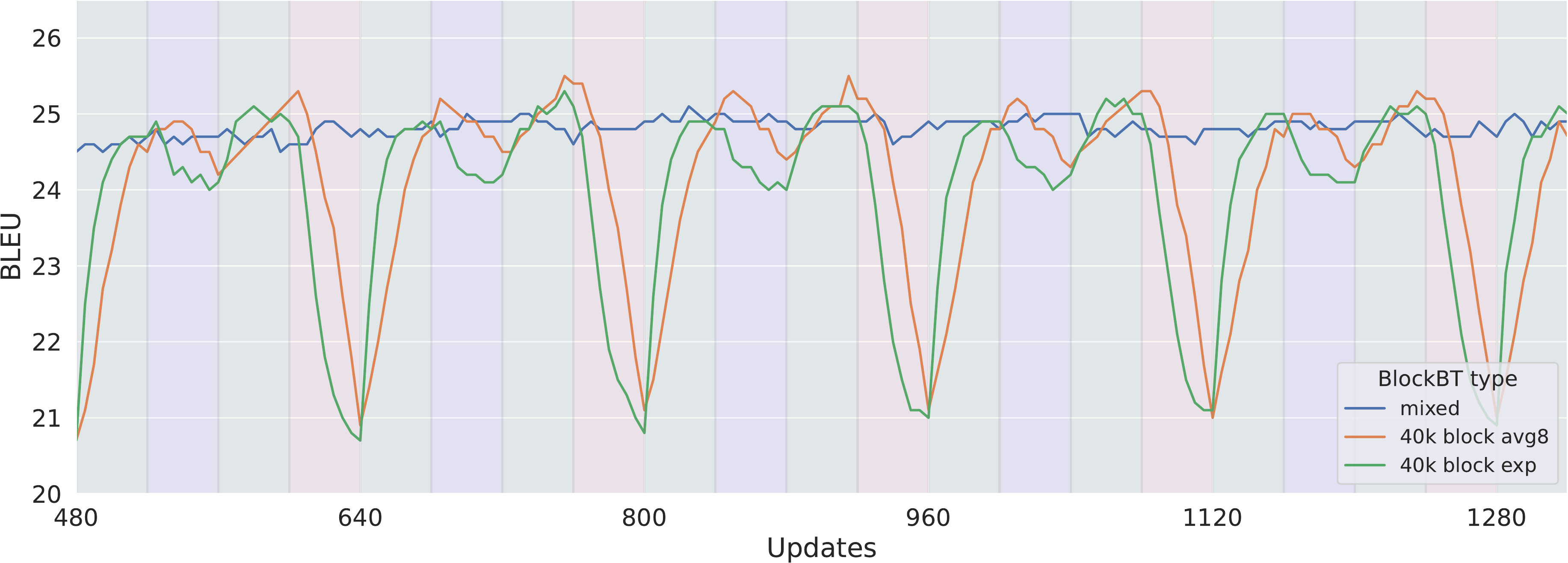}
\caption{Comparison of different training regimes for CS$\rightarrow$EN translation on \texttt{newstest2020} in terms of BLEU (top) and COMET (bottom). Background colors for block-BT regime show which part of training data was used for a given part of the training. Green means authentic parallel data, blue is CS$\rightarrow$EN forward translation and red is EN$\rightarrow$CS backtranslation.}
\label{fig:mixed_avg_exp_czen}
\end{figure*}

\begin{table}[p]
\scriptsize
{
\begin{tabular}{l@{~~~}r@{~~~}r@{~~~}r@{~~~}r@{~~~}r@{~~~}r@{~~~}r}\toprule
               &               &                  &\multicolumn{2}{c}{\textbf{best BLEU}} &\multicolumn{2}{c}{\textbf{best COMET}} \\
\textbf{Model} &\textbf{Block} &\textbf{Avg type} &\textbf{update (k)} &\textbf{BLEU} &\textbf{update (k)} &\textbf{COMET} \\
\midrule
\multirow{6}{*}{base} &\multirow{2}{*}{mixed} &exp &1405 &25.2 & 1220 &0.4149  \\
& &exp+avg8 &1430 &25.1 & 1220 &0.4114  \\
&\multirow{4}{*}{40k} &- &580 &25.3 & 1040 &0.4086  \\
& &exp &755 &25.3 & 570 &0.4183  \\
& &avg8 &765 &25.4  &1060 &0.4175  \\
& &exp+avg8 &1080 &25.2 & 1230 &0.4186  \\

\bottomrule
\end{tabular}}
\caption{COMET and BLEU scores for Czech to English directions. The best checkpoints were chosen based on their performance on \texttt{newstest2020}. }\label{tab:csen_results}
\end{table}
\paragraph{Reverse direction} For the reverse direction, Czech to English, we performed less extensive evaluation. We only compare \textit{mixed}, \textit{block-BT} with 40k blocks and either exponential smoothing or checkpoint averaging. The behavior of the metrics is shown in Figure \ref{fig:mixed_avg_exp_czen} and the final best scores on \texttt{newstest2020} are presented in Table \ref{tab:csen_results}. \textit{Block-BT} still outperforms \textit{mixed} training, but by a smaller margin than in the other direction. We attribute this difference to the fact that the Czech side of the CzEng dataset is more often translationese (originally English text translated into Czech) and thus differs more from \textit{csmono} part, giving space for the larger gains.

\paragraph{Backtranslation direction}

\begin{table}[!htp]\centering
\scriptsize
\begin{tabular}{lrrrrrrr}\toprule
    &       &         &\multicolumn{2}{c}{\textbf{BLEU}} &\multicolumn{2}{c}{\textbf{COMET}} \\
\bf dir & \bf regime & \bf datasets & \bf Dev & \bf Test & \bf Dev & \bf Test \\\midrule
\multirow{6}{*}{encs} &\multirow{3}{*}{mixed} &all &34.7 & 20.9 &0.7337 & 0.6206 \\
& &auth+cs & 31.5 & 19.5 & 0.6904 & 0.5779\\
& &auth+en & 34.8 &20.6 &0.7258 &0.6097 \\
&\multirow{3}{*}{block} &all &35.3 & \textbf{21.1 }&0.7474 & \textbf{0.6245}\\
& &auth+cs &33.9 & 19.9&0.7232 & 0.5908\\
& &auth+en & \textbf{35.4} & 20.7 &\textbf{0.7497 }& 0.6147 \\ \midrule
\multirow{3}{*}{csen} &\multirow{1}{*}{mixed} &all & 25.2& - &0.4149 & - \\
&\multirow{2}{*}{block} &all & 25.3 & - &0.4183 & - \\
& &auth+en & 24.3& - &0.3682 & - \\
\bottomrule
\end{tabular}
\caption{Results on newstest2020 and newstest2021 for various dataset combinations on dev (\textit{newstest2020}) and test (\textit{newstest2021}) sets, respectivelly, COMET scores are computed by wmt20-comet-da model.}\label{tab:bt_direction}

\end{table}

We also evaluate influence of using only backtranslations (i.e. \textit{csmono} for en$\rightarrow$cs) as additional synthetic data  (monolingual data in target language  automatically translated to source language) or adding also forward translations (automatic translations from source language to target; \textit{enmono}) and we present the results in Table \ref{tab:bt_direction}. Interestingly, the results show large gains in both BLEU and COMET when using forward translation. We hypothesize this is caused by the good quality of the model used to create the translation for \textit{enmono}. In such case, the translation model plays the role of the teacher in teacher$\rightarrow$student training and might lead to good quality results.

\paragraph{Named entities test sets}

\begin{figure*}[h]
\centering
\includegraphics[width=0.95650\textwidth]{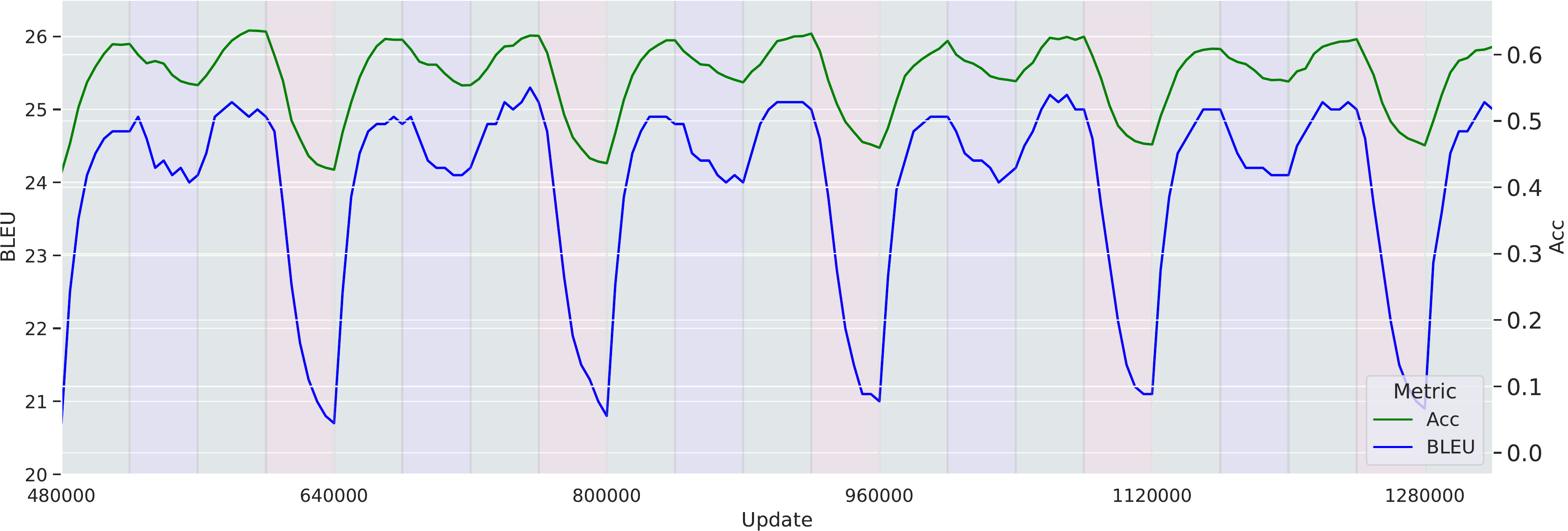}

\includegraphics[width=0.95650\textwidth]{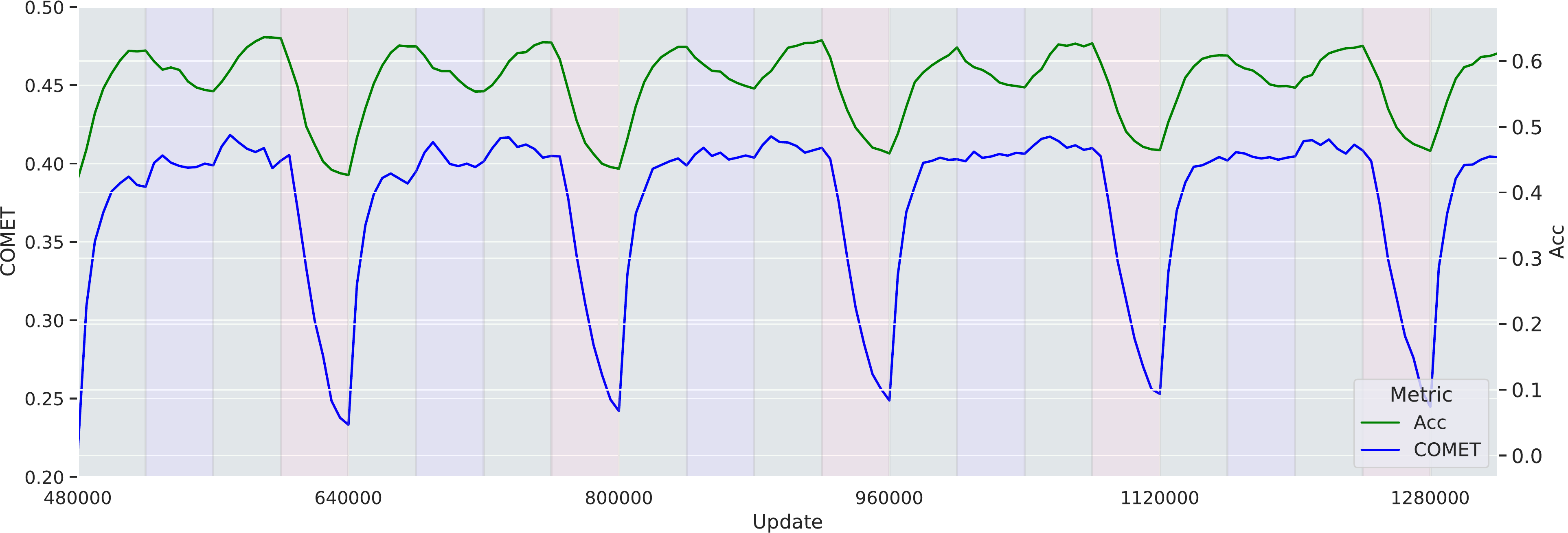}

\caption{Behaviour of BLEU (top), COMET (bottom) on \texttt{newstest2020} and NE translation accuracy on \texttt{restaurant} test set for Czech to English translation with block-BT using exponential smoothing.}
\label{fig:ne_acc}
\end{figure*}

\begin{table}[!htp]\centering
\small
\begin{tabular}{lrrrr}\toprule
\textbf{Update (k)} &\textbf{COMET} &\textbf{BLEU}  &\textbf{Acc}\\\midrule
570  &\textbf{0.4183} &24.9 &0.607\\
755&0.4038 &\textbf{25.3}  &0.629  \\
590  &0.4099 &24.9 & \textbf{0.636} \\
\bottomrule
\end{tabular}
\caption{Best checkpoints of Czech to English model trained with 40k blocks and exponential smoothing in terms of COMET (first row), BLEU (second row) on newstest2020 and NE translation accuracy on restaurant test set (third row).}
\label{tab:ne_acc}
\end{table}

From anecdotal evidence, we have seen that checkpoints with large influence of backtranslated data perform worse on named entities (NE) translation and COMET and BLEU scores might not reflect this drop of accuracy. We evaluate the models in terms of accuraccy of NE translation on the \texttt{restaurant} test set.
We selected Czech to English direction, since the evaluation is easier given lower morphological richness of the target language. \todo[]{I also have results for en-cs, I should integrate them}Figure  \ref{fig:ne_acc}  shows comparison of behavior of NE (NE) translation accuracy on the restaurant test set and COMET and BLEU scores on \texttt{newstest2020} for exponential smoothing and checkpoint averaging. NE accuracy peaks towards the end of \textit{auth} regions (green). Both COMET and BLEU scores peak also during the \textit{auth} part of the training, but, especially for COMET, the peak occurs in earlier stages after the switch to \textit{auth}. Overall, BLEU curve correlates better with the NE accuracy curve. We hypothesize this might be related to the fact that COMET was found to be insensitive to NE errors by \citet{amrhein2022identifying}.

However, it seems that the shift between the accuracy and the other two metrics is not too large in our settings and choosing the best performing model in terms of either COMET or BLEU should not hurt NE translation by a large amount. We further investigate that in Table \ref{tab:ne_acc} -- we chose the checkpoint with the best COMET (first row) and the best BLEU (second row)  on the \texttt{newstest2020} and the checkpoint with the best NE translation accuracy on the restaurant test set (third row). We compute all three metrics for these three models. The best COMET checkpoint obtains accuracy of 60.7\% on the restaurant test set, the best BLEU checkpoint reaches the accuracy of 62.9\%, while the best accuracy reached by any checkpoint is 63.6\%. 

\todo{Do this for noexp avg, since it seems it also peaks in csmono regions which could kill the acc}
\todo{So for en->cz, the ne acc is actually lowest in auth, while for cz->en, it is highest in auth? Weird, check brno results also}


\subsection{MBR reranking}

\begin{table}[!htp]\centering
\scriptsize
\begin{tabular}{lrrrrrr}\toprule
\textbf{i} &\textbf{auth} &\textbf{cs} &\textbf{en} &\textbf{AVG comet20} &\textbf{MBR comet20} &\textbf{comet21} \\\midrule
1 & - &- &- &0.7322 & 0.7888 & 0.0885 \\
2& 9 &2 &1 &\textbf{0.7430} &0.8082 &0.0946 \\
3&4 &4 &4 &0.7408 &0.8182 &0.0972 \\
4&12 &0 &0 &0.7425 &0.8010 &0.0929 \\
5&0 &12 &0 &0.7303 &0.8104 &0.0949 \\
6&0 &0 &12 &0.7372 &0.7960 &0.0918 \\
7&1 &7 &4 &0.7370 &\textbf{0.8232} &\textbf{0.0981} \\
8&0 &7 &5 &0.7361 &\textbf{0.8232} &\textbf{0.0980 }\\
9&2 &7 &3 &0.7377 &\textbf{0.8231} &\textbf{0.0981 }\\
\bottomrule
\end{tabular}
\caption{Results of MBR reranking on \texttt{newstest2020} for different selection of the hypotheses n-best lists produced by checkpoints from different training blocks. In total, 12 n-best lists produced by transformer-base models are concatenated and the first three columns show how many n-best lists are used from each block (the checkpoints for each block are sorted by COMET (wmt20-da model), so these are produced by the best performing checkpoints). The \textit{AVG COMET20} shows the average  wmt20-da COMET scores for the first hypotheses of each n-best list that was used,\textit{ MBR COMET20} shows  wmt20-da score of the final sentences after MBR reranking,  COMET21 shows results of the same sentences from wmt21-da model.}\label{tab:mbr}

\end{table}

We used MBR reranking to rerank concatenation of n-best lists produced by various checkpoints. In total, we used 6-best lists from 12 checkpoints, i.e. 72 candidate hypotheses for each sentence. We divided the checkpoints based on which block of the training data they were saved in and sorted them by COMET score on \texttt{newstest2020}. Using different strategies we selected the best performing checkpoints to provide the n-best lists.  We present the results in Table \ref{tab:mbr}. The first row shows results for mixed-BT regime, i.e. we concatenated n-best lists produced by the 12 best performing mixed-BT checkpoints. In the second row, the block-BT training checkpoints were used to create n-best lists, selected only based on their COMET scores, without any regard on the block type they were saved in. In third row, we combine n-best lists from 4 best checkpoints from each type of block. In rows 4-6, we use best checkpoints from each type of block separately. In the final three rows, we show the optimal selections which yielded the best score. The results suggest that larger diversity across block types of the checkpoints improves MBR results: the combination of n-best lists produced by checkpoints from diverse block types provides better hypotheses for MBR, even though the average COMET score of these checkpoints is lower than for the less diverse selection (see rows 2 and 3).

\todo[]{Todo: I should use BLEURT for final evaluation, since I used comet as a utility function and I want to see if other metrics also improve}
\todo[]{I could use dictionary and constrained model to translate rare words (with all the possible dict translations) and add these translations into the pool, but probably in some other paper, it would be too much for a system description, the problem is that comet is not very sensitive to rare word translation}
\todo[]{I could use combination of different metrics for mbr, but i need to scale them properly}

\subsection{Submission}
\begin{table}[h]\centering
\footnotesize
\begin{tabular}{l@{~~}r@{~~}r@{~~}r@{~~}r@{~~}r}\toprule
\textbf{auth} &\textbf{cs} &\textbf{en} &\textbf{AVG comet20} &\textbf{MBR comet20} &\textbf{comet21} \\\midrule
 9 &2 &8 &0.7802  & 0.8566 & 0.1114 \\
\bottomrule
\end{tabular}
\caption{Our final submission for the EN-CS general translation task, based on outputs of the transformer-big 12-6 model. Meaning of the columns is identical to Table \ref{tab:mbr}.}\label{tab:final}

\end{table}
\begin{table}[h]\centering
\footnotesize
\begin{tabular}{lr@{~~~}r@{~~~}r@{~~~}r}\toprule
System&\llap{COMET-B} &COMET-C &ChrF-all \\\midrule
Online-W &97.8 &79.3 &70.4 \\
Online-B &97.5 &76.6 &71.3 \\
CUNI-Bergamot * &\textbf{96.0 }&\textbf{79.0} &65.1 \\
JDExploreAcademy * &95.3 &77.8 &\textbf{67.2} \\
Lan-Bridge &94.7 &73.8 &70.4 \\
Online-A &92.2 &71.1 &67.5 \\
CUNI-DocTransformer * &91.7 &72.2 &66.0 \\
CUNI-Transformer * &86.6 &68.6 &64.2 \\
Online-Y &83.7 &62.3 &64.5 \\
Online-G &82.3 &61.5 &64.6 \\
\bottomrule
\end{tabular}
\caption{Results of automatic metrics on WMT22 General Task test set. Constrained submissions are marked by an asterisk, the best scores among constrained submissions are bold. COMET-B and COMET-C are COMET scores for the two different references, ChrF is computed using both references together.}\label{tab:test_set}

\end{table}
Our primary submission is based on the \textit{big 12-6} model and MBR reranking. We explored all the possible combinations of 18 checkpoints from different blocks as described in the previous section. The results of the best combination are shown in Table \ref{tab:final}.
We present the results of the official evaluation in our task in Table \ref{tab:test_set}. There were 5 submitted systems (4 constrained) and 5 online services. Our submission ranks first in COMET score among the constrained systems and third in ChrF.
\section{Conclusion}
We describe our submission to WMT22 and experiments that have led to development of our system. We confirm effectiveness of block-BT on the recent COMET metric. We demonstrate the behavior of the translation quality over the course of the training and discuss the effects of various settings. We also show that MBR reranking can benefit from more diverse checkpoints created by block-BT training. 

\section{Acknowledgements}
\small
This work was supported by GAČR EXPRO grant NEUREM3 (19-26934X), Bergamot project (European Union’s Horizon 2020 research and innovation programme under grant agreement No 825303) and by the Ministry of Education, Youth and Sports of the Czech Republic, Project No. LM2018101 LINDAT/CLARIAH-CZ.

\FloatBarrier
\todo[]{compute bleurt for best single and then for the mbr rescored}
\bibliography{anthology,custom}
\bibliographystyle{acl_natbib}

\end{document}